\documentclass[conference]{IEEEtran}
\IEEEoverridecommandlockouts

\usepackage[square,numbers]{natbib}

\usepackage{url} 

\usepackage[margin=0.97in]{geometry} 
\usepackage[dvips]{graphicx}
\usepackage{graphicx}
\usepackage{float}
\usepackage{amsmath}
\usepackage{amssymb}
\usepackage{mathrsfs}
\usepackage{amsfonts}
\usepackage{color}
\usepackage{physics}
\usepackage{amsmath, amssymb}
\usepackage{booktabs} 
\usepackage{multirow}
\usepackage{multicol}
\usepackage{threeparttable}

\usepackage{amsmath,amssymb,amsfonts}
\usepackage{graphicx}
\usepackage{booktabs}
\usepackage{textcomp}
\usepackage{multirow}
\usepackage{algpseudocode}
\usepackage{amsmath}

\usepackage[plain,noend,ruled,linesnumbered]{algorithm2e}
\SetKwProg{Init}{initial}{}{}

\begin{document}
\pagenumbering{arabic}
\pagenumbering{gobble}

%
\title{Interpretable Data Fusion for Distributed Learning: A Representative Approach via Gradient Matching\\
\thanks{B. Geng and K. Li are the corresponding authors.}
}
\author{\IEEEauthorblockN{Mengchen Fan$^{a}$, Baocheng Geng$^{a}$, Keren Li$^{b}$, Xueqian Wang$^c$ and Pramod K. Varshney$^d$}
\IEEEauthorblockA{$^a$\textit{Department of Computer Science, University of Alabama at Birmingham, Birmingham, US}\\
$^b$\textit{Department of Mathematics, University of Alabama at Birmingham, Birmingham, US}\\
$^c$\textit{Department of Electronic Engineering, Tsinghua University, Beijing, China}\\
$^d$\textit{Department of Electrical Engineering and Computer Science, Syracuse University, Syracuse, US}
}
}

\maketitle

\begin{abstract}

This paper introduces a representative-based approach for distributed learning that transforms multiple raw data points into a virtual representation. 
Unlike traditional distributed learning methods such as Federated Learning, which do not offer human interpretability, our method makes complex machine learning processes accessible and comprehensible. It achieves this by condensing extensive datasets into digestible formats, thus fostering intuitive human-machine interactions. Additionally, this approach maintains privacy and communication efficiency, and it matches the training performance of models using raw data.
Simulation results show that our approach is competitive with or outperforms traditional Federated Learning in accuracy and convergence, especially in scenarios with complex models and a higher number of clients. This framework marks a step forward in integrating human intuition with machine intelligence, which potentially enhances human-machine learning interfaces and collaborative efforts.

\end{abstract}

\begin{IEEEkeywords}
Data fusion, distributed learning, interpretability, human integrated AI.
\end{IEEEkeywords}

\section{Introduction}

Fusion of information from diverse sources, such as sensors, provides a richer, more comprehensive perspective on a phenomenon of interest, thereby enhancing predictive accuracy and decision-making efficiency \cite{varshney2012distributed,veeravalli2012distributed,kailkhura2015distributed,zhang2019fusion,quan2023ordered}. In numerous application scenarios, data are gathered and maintained locally by various agents or facilities. Recently, the imposition of stricter data privacy regulations and limitations on data communication bandwidth have driven the development of distributed learning systems capable of training a centralized model without the need to exchange raw data.

One notable approach, Federated Learning (FL)  allows for the exchange of model parameter updates instead of raw data, thus preserving privacy, minimizing data transmission, and reducing the reliance on centralized data repositories. A variety of FL algorithms have been devised to tackle different machine learning (ML) tasks, including deep neural networks (DNNs) \cite{assran2019stochastic,liu2019communication,li2020federated}, gradient boosted decision trees (GBDTs) \cite{cheng2021secureboost,li2020practical}, logistic regression \cite{chen2018privacy}, and support vector machines (SVMs) \cite{smith2017federated}. These architectures are tailored to meet specific requirements such as data distribution, communication limits, and privacy needs in distributed, decentralized, and hierarchical structures \cite{mcmahan2017communication,stich2018local,lian2017can,koloskova2020unified,quan2023efficient,quan2021strategic,quan2022enhanced}.



In the design and implementation of these ML systems, a major challenge is the effective integration of human insights into decision-making and oversight. Traditionally, the emphasis has been solely on the accuracy of ML models, neglecting the need for comprehensibility. This approach does not recognize the collaborative essence of such systems in critical environments, where decisions are made through the combined efforts of humans and ML models. In scenarios where small details have significant impacts on outcomes, exclusive reliance on automated decision-making systems is not advisable \cite{geng2022human}.


Several studies within the signal processing and ML literature have investigated collaborative human machine decision-making, focusing on how humans interpret and perceive data and ML outputs \cite{geng2022human,geng2021collaborative,geng2021cognitive,8976222,geng2020prospect}. Yet, these investigations frequently neglect a vital consideration: the ML models must be designed to be interpretable, ensuring that they can interact with humans at a reasonable pace and in an easily understandable format.
Many ML models are black boxes that offer little explanation of their training and inference processes in terms of understandability to humans. This is particularly true in FL, where the model parameter updates from each client lack intuitive physical meaning and are incomprehensible to human operators. For example, while humans can easily interpret an image, understanding the model parameter updates in a Deep Neural Network (DNN) system is far from straightforward.



Inspired by the score-matching representative (SMR) proposed in \citep{li2022smr}, we present a distributed learning framework centered around the concept of representative. A `representative data point', or simply `{\it representative}' is defined as a synthesized data point derived from each data block on decentralized devices. This representative is computed to have the same effect as the original raw
dataset on the model's weight parameters. Specifically, SMR is designed to replicate the gradient of the loss function on the model parameters, ensuring that its impact on model updates mirrors that of the raw data.

There have been previous studies \cite{zhao2020dataset, domke2012generic, cazenavette2022dataset} utilizing gradient matching techniques that have demonstrated the ability to compress large datasets into smaller, manageable subsets. 
However, this process is accompanied by a certain degree of loss in model performance. Our method extends beyond mere dataset reduction to model training in distributed environments in real-time. Moreover,  it ensures the training accuracy of using a representative is closer to that of the original data by introducing a residual parameter.  Additionally, our approach enhances the interpretability of the representative, by pertubating a small noise based on the mean of the dataset, so that representatives are understandable to humans. Also, our approach favors a real-time decision-making method that enhances efficiency, responsiveness, and accuracy by enabling immediate actions based on current data in the distribution environment.

Our framework aims to ensure that employing a single representative data point achieves outcomes comparable to those derived from using the raw data set within the block. This approach offers several advantages:

\begin{itemize}
    \item Reduced Communication Overhead: By condensing large datasets into a single representative, our method lowers the communication burden between the clients and the servers.
    \item Privacy Preservation: The representative is a virtually constructed data point that encapsulates aggregated data information, thus protecting individual-specific details and ensuring privacy.
    \item Enhanced Interpretability: Unlike FL, where only model parameter updates are shared, our approach maintains data's original structure and format through the representative, enhancing interpretability for human operators. This feature is crucial for systems requiring human-centric decision-making.
    \item Improved Performance: Numerical results show that our framework surpasses the FL approach, specifically FedAVG, in accuracy metrics. In addition, our method demonstrates smoother and more rapid convergence, highlighting its efficiency and effectiveness.
\end{itemize}



\section{Problem Formulation}



In distributed learning tasks, we define a dataset distribution $\mathscr{D}$ composed of $n$ data points, represented as $x = (x_1, \ldots, x_n)^T$, where each $x_i \in \mathbb{R}^d$, and the corresponding system outputs (or labels) as ${y} = (y_1, \ldots, y_n)^T$ for $i=1,\ldots, n$. The task of learning and inference in this context can be articulated as an optimization problem:

\begin{equation}\label{obj}
\min_{\mathrm{w}} L(\mathrm{w}; x,y) = l(\mathrm{w}; x,y) + r(\mathrm{w}; \lambda).
\end{equation}

Here, $L$ symbolizes the total loss of the system, with $\mathrm{w}$ representing the model parameters to be estimated. The term $l(\mathrm{w}; x,y)$ denotes a loss function or the negative of a log-likelihood function  \cite{mccullagh1989glm}, and $r(\mathrm{w}; \lambda)$ denotes the regularization term, where $\lambda$ is the tuning parameter.

Considering $\mathscr{D}$ as a dataset dispersed across $K$ parallel clients, we introduce $I = {1, 2, \ldots, n}$ as the complete set of data indices. The subset of data on the $k$-th client is given by $\mathscr{D}_k = \{(x_i,y_i) | i \in I_k\}$, with $I_k \subseteq I$ denoting the indices subset associated with node $k$. The size of each node, $n_k$, is determined by the number of indices in $I_k$. The collection of indices ${I_1, I_2, \ldots, I_K}$ effectively partitions $I$.
A fusion center (or central server) $\mathcal{C}$, is employed to facilitate the model training process.


Li and Yang (2022) \citep{li2022smr} examined a scenario in which the loss function corresponds to a convex objective function derived from the likelihood of a generalized linear model (GLM). The score-matching representative (SMR) method enables each client/node to compute its representative data by aligning the score function—specifically, the gradient of the log-likelihood function $l$ in \eqref{obj} with respect to $\mathrm{w}$—with the client's raw dataset \cite{mccullagh1989glm}. This approach demonstrated that the SMR-based estimator $\tilde{\mathrm{w}}^{SMR}$ achieves accuracy equivalent to the ground truth estimator $\hat{\mathrm{w}}$. Furthermore, it was established that $\tilde{\mathrm{w}}^{SMR}$ and $\hat{\mathrm{w}}$ differ by an order of magnitude proportional to $\sqrt{\Delta}$, where $\Delta = \max_k\max_{i,j} \norm{x_i - {x_j}}_2$ represents the maximum Euclidean distance (L2 norm) between any two data points within the dataset.

In the next section, we present a significant extension to our analysis to include general non-convex functions, particularly focusing on DNNs. 
Our goal is to enhance the effectiveness and applicability of representation learning in artificial intelligence by tackling the challenges of non-convex optimization problems.
To achieve this goal, we will introduce innovative frameworks and techniques tailored specifically to accommodate the complexities inherent in non-convex optimization landscapes. 
By addressing these challenges, we seek to equip representation learning with the flexibility and strength necessary to solve various optimization and learning tasks. These efforts aim to transform the field of distributed machine learning and push the boundaries of AI research forward.

\section{Methodology}


In machine learning tasks, especially within DNNs utilized for a variety of standard classification tasks, the objective function often exhibits complicated non-convex characteristics. Consider an input feature vector $x_i \in \mathbb{R}^d$ for data sample $i$, and a corresponding label $y_i \in [M]$, where $M$ denotes the number of potential classes. A commonly employed loss function is the cross-entropy loss. For example, when $M=2$, the cross-entropy loss is defined as $l(\mathrm{w}, x, y) = -\sum_{i=1}^{N} y_i \log(f(\mathrm{w}, x_i))$, in which $y_i$ represents the true label distribution for the input $x_i$, $\mathrm{w}$ denotes the model parameters, and $f$ symbolizes the model function. In this analytical exploration, we disregard the regularization term in \eqref{obj}, thereby having $L(\mathrm{w}, x, y)=l(\mathrm{w}, x, y)$.

Optimization of this expected loss typically employs gradient-based algorithms such as Stochastic Gradient Descent (SGD) or Adam. These methods iteratively update the parameters $\mathrm{w}$ in the direction that reduces the expected loss, following the update rule $\mathrm{w}_{\text{new}} = \mathrm{w}_{\text{old}} - \eta \nabla_\mathrm{w} L(\mathrm{w}, x, y)$, where $\eta$ represents the learning rate.

\subsection{Representative-based Centralized Learning}


We begin by exploring the representative-based centralized model training framework, utilizing the dataset $\mathscr{D}$. At each epoch, rather than selecting a batch of data of size $B$ from $\mathscr{D}$, denoted as $\mathcal{B} = \{(x_i, y_i) \mid i \in I_{\mathcal{B}}\}$, 
to train the model $f$, we compute a representative $({x^r}, y^r)$ such that training the model with the representative\footnote{The representative $(x^r, y^r)$ preserves the data dimensionality of the raw data points, which facilitates its interpretability and comprehensibility for humans.} induces the same parameter updates as would training with the entire batch $\mathcal{B}$. This section presents key algorithms for constructing the representative that will be applied in distributed learning in Section III.B.

To ensure that a single virtual point can effectively represent the data within a batch, all data points in $\mathcal{B}$ are supposed to share the same label, such that $y_i = y^{\mathcal{B}}$ for all $i \in I_{\mathcal{B}}$. Hence, we set the representative label $y^r$ equal to $y^{\mathcal{B}}$.

Let $\nabla_{\mathrm{w}} L(\mathcal{B}) = \sum_{i \in I_{\mathcal{B}}}\nabla_{\mathrm{w}} L(\mathrm{w}, x_i, y_i)$ represent the gradient with respect to $\mathrm{w}$ for the current model $f$ based on training $\mathcal{B}$. Our objective is to determine $x^r \in \mathbb{R}^d$ such that the representative $(x^r, y^r)$ generates a gradient on $\mathrm{w}$ for $f$ that matches the mean gradient produced by the data points in $\mathcal{B}$, $\nabla_{\mathrm{w}} L(\mathcal{B})/B$. The optimization problem becomes searching for $x^r$ that minimizes the representative loss $l^r$:
\begin{equation}
    l^r = \norm{\nabla_\mathrm{w}L(\mathrm{w}, x^r, y^r) - \frac{\nabla_{\mathrm{w}} L(\mathcal{B})}{B}}_2.
\end{equation}


To ensure that the representative $x^r$ remains close to the original data points in batch $\mathcal{B}$, thereby further preserving intuitive insight into the data, we search for $x^r$ by adding a certain magnitude of perturbations $\delta$ to the mean of the data points, denoted as $\Bar{x} = \frac{1}{B}\sum_{i\in I_{\mathcal{B}}}x_i$. This process is similar to adversarial training, with the distinction that our goal is to identify a perturbation $\delta^*$ that adheres to a magnitude constraint (e.g., $|\delta|_2 \leq s$ for a given norm $p$ and a specified constant $s$) to minimize the loss $l^r$. The optimization problem is thus formulated as:

\begin{equation}\label{deltastar}
\delta^* = \underset{\norm{\delta}_2\leq s}{\arg\min} \norm{\nabla_{\mathrm{w}}L(\mathrm{w}, \Bar{x}+\delta, y^r) - \frac{\nabla_{\mathrm{w}} L(\mathcal{B})}{B}}_2,
\end{equation}
where $\delta^*$ is determined through gradient descent for a total $H$ epochs. Ultimately, we set $x^r = \Bar{x} + \delta^*$, and effectively construct $(x^r,y^r)$ as the representative that closely aligns with the batch's original data while also minimizing the targeted loss function.


Once the representative point $(x^r, y^r)$ is determined, we train the model using this synthesized point instead of the raw data from batch $\mathcal{B}$. The gradient with respect to the model parameters at this point is denoted as $\nabla \mathrm{w}^r = \nabla_{\mathrm{w}} L(\mathrm{w}, x^r, y^r)$. Model parameters are then updated based on the gradient $B \nabla \mathrm{w}^r$ for the current epoch.


Furthermore, we acknowledge the challenge of achieving an exact match between the gradient produced by the representative and that generated by the full data batch. There exists a discrepancy between $B \nabla \mathrm{w}^r$ and $\nabla_{\mathrm{w}} L(\mathcal{B})$. To address this, we introduce an error residual term $\tau_i$ to account for the difference between these two gradient terms. This allows us to integrate the residual error into the computation of $x^r$ at each round of training. Consequently, the optimization problem for determining $\delta^*$ is updated as follows:

\begin{equation}\label{deltawithresidual}
\delta^* = \underset{\norm{\delta}_2\leq s}{\arg\min} \norm{\nabla_\mathrm{w}L(\mathrm{w}, \Bar{x}+\delta, y^r) - \frac{\nabla_{\mathrm{w}} L(\mathcal{B})}{B} + \tau}_2,
\end{equation}
where $\tau = \nabla \mathrm{w}^r - \nabla_{\mathrm{w}} L(\mathcal{B})/B$ represents the residual error accrued in the previous epoch. Simulation results presented in Section IV demonstrate that incorporating the residual error significantly enhances the efficacy of representative-based training.

The overall optimization process in this representative-based central learning framework is summarized as follows:
\begin{equation}
\min_{\mathrm{w}} \mathbb{E}_{ \mathcal{B}\sim \mathcal{D}} L(\mathrm{w}, x^r, y^r),
\end{equation}
with $x^r,y^r$ denoting the representative of batch $\mathcal{B}$, where $x^r = \Bar{x} + \delta^*$, and $\delta^*$ is derived according to \eqref{deltastar} at every epoch. The details of this proposed framework
are shown in Algorithm \ref{alg:1}.

\begin{algorithm}[htb]
\label{alg:1}
\caption{Representative-based Centralized Learning}
\SetAlgoLined
\KwIn{Initialize model parameters $\mathrm{w}$, dataset $\mathcal{D}$, batch size $B$, learning rate $\eta_{\mathrm{w}}$}
\KwOut{Trained model}
Initialize residual term $\tau = 0$;
\For{$t=1$ \KwTo $T$}{
    Sample a batch of data $\mathcal{B}$ with same label $y^{\mathcal{B}}$;

    Compute the gradient of the raw data in $\mathcal{B}$ on model parameters, denoted by $\nabla_{\mathrm{w}} L(\mathcal{B})$;
    
    Compute the representative of the batch $(x^r,y^r)$ such that $x^r = \Bar{x}+\delta^*$ as in \eqref{deltawithresidual} and $y^r = y^{\mathcal{B}}$;

    Update model parameters using the gradient computed from $(x^r,y^r)$: $\mathrm{w} = \mathrm{w} - \eta_{\mathrm{w}} B \nabla \mathrm{w}^r$;

    Update the residual term: $\tau =\nabla \mathrm{w}^r - \nabla_{\mathrm{w}} L(\mathcal{B})/B $;
}
{\bf{return}} $\mathrm{w}$
\end{algorithm}

It is important to highlight that synthesizing all data points in batch $\mathcal{B}$ into a single virtual representative simplifies the visualization of the training process for humans. In each training epoch, observers need only focus on a single data point rather than $B$ individual points. This significantly aids human oversight by facilitating the detection of outliers and enhancing the comprehension of the training dynamics, among other benefits.

\subsection{Representative Data Fusion for Distributed Learning }

In the distributed learning paradigm, datasets are dispersed across $K$ clients. This setup reduces to the centralized training as a special case when $K=1$. Traditionally, Federated Learning (FL) techniques, particularly the FedAVG \cite{mcmahan2017communication} serving as the baseline comparison model in this paper, have been employed to avoid raw data transmission and preserve privacy. In this section, we introduce a representative-based distributed learning strategy that not only preserves the benefits but also enhances human comprehension of the training data. Moreover, this method has the potential to exceed FL techniques in convergence speed and accuracy.

Consider a scenario where the total dataset $\mathcal{D}$ is distributed among $K$ clients, with each client possessing a subset of the dataset, denoted by $\mathcal{D}_k$. Similar to the FL methodology, at the beginning of each epoch, the server broadcasts the global model parameters $\mathrm{w}$ to all clients. Subsequently, each client randomly samples a batch of data points that share the same label. Then, a representative $(x^k, y^k)$ is computed based on the data points in the batch and the global model. This representative, along with the batch size $B_k$, is then transmitted back to the server. Utilizing these representatives, the server updates the global model. In particular, each client maintains a residual term to correct the gradient discrepancy between its representative and the entirety of its batch data. This procedure is iterated over $T$ rounds to refine the global model.

For the sake of clarity, we assume a uniform batch size $B$ across all clients. This constraint can be adapted to heterogeneous batch sizes to reflect real-world application scenarios. The steps of this representative-based distributed learning strategy are shown in Algorithm \ref{alg:2}.

\begin{algorithm}
\label{alg:2}
\SetAlgoLined
\KwIn{Model parameters $\mathrm{w}$, batch size $B$, number of clients $K$, learning rate $\eta_{\mathrm{w}}$, global training epoch $T$ and local training epoch $H$}
\KwOut{Globally trained model}
\textbf{Server executes:}\\
Initialize global model parameters $\mathrm{w}_0$\;
\For{$t=1$ \textbf{to} $T$}{
  Select a subset of clients $S_t$ for training\;
  \For{each client $k \in S_t$ {in parallel}}{
    Each client updates its representative: $(x^k_t,y^k_t) \leftarrow$ \textbf{ClientRep}$(k, \mathrm{w}_{t-1}, \tau^k_{t-1})$\;
  
  $\mathrm{w} \leftarrow \mathrm{w} - \eta_{\mathrm{w}} B\nabla_{\mathrm{w}} L(\mathrm{w}, x^k_t,y^k_t)$\;}

}
\KwRet $\mathrm{w}$\;
\textbf{ClientRep}$(k, \mathrm{w}, \tau)$:\\
\Indp
Sample a batch of data $\mathcal{B}$ with same label $y^{\mathcal{B}}$;

    Compute the gradient of the raw data in $\mathcal{B}$ on model parameters, denoted by $\nabla_{\mathrm{w}} L(\mathcal{B})^k$;
    
    Each client learns $\delta^*$ as in \eqref{deltawithresidual} using $H$ epochs and compute its representative $(x^k,y^k)$  such that $x^k = \Bar{x}+\delta^*$ and set $y^k = y^{\mathcal{B}}$;

    Update its residual term: $\tau^k =\nabla_{\mathrm{w}} L(\mathrm{w},x^k,y^k) - \nabla_{\mathrm{w}} L(\mathcal{B})^k/B $;

  \KwRet $(x^k,y^k)$\;
\caption{Representative-based Distributed Learning with Server and Multiple  Clients}
\end{algorithm}

{\it Computational Complexity:}
The computational complexity in representative-based distributed learning is influenced by three main factors: 1) Participation Proportion 
$P$: This factor measures the percentage of clients involved in each computation round, indicating the extent of collaborative efforts utilized for model training across distributed clients at each stage.
2) Local Iterations 
$H$: This represents the number of training iterations performed by each client on its local dataset to calculate its representative in every round. 
$H$ indicates the computational efforts towards ensuring the representative's impact closely mirrors that of the raw data.
3) Batch Size 
$B$: This parameter specifies the size of the local batch used for generating the representative. It is a measure of the level of abstractness in data representation.
Exploring the trade-off among these parameters to discern their collective impact on system performance is a potential research direction. Given the page constraints, these considerations are reserved for future investigation.


\section{Simulation Results}

\subsection{Experimental Setup}


In our experiments, we assess our methodology using two datasets. The first is the Image Segmentation dataset \cite{misc_image_segmentation_50}, a compact multivariate dataset comprising image data characterized by high-level, numeric-valued attributes. The second dataset is MNIST, a standard benchmark featuring 60,000 images of handwritten digits. 
{For the distributed learning problem considered here, we adopt a non-Independent and Identically Distributed (non-IID) strategy, as discussed in \cite{zhao2018federated,hsu2019measuring}, to partition the dataset across various clients, ensuring each client contains data with the same label, thus representing non-IID data characteristics. Specifically, we allocate 200 images with the same label to each client, mirroring real-world scenarios of non-IID data distribution.}
We examine the performance of two neural network architectures: a Multilayer Perceptron (MLP) model, which consists of several layers with ReLu activation functions; and a Convolutional Neural Network (CNN) with two convolutional layers and one linear layer, which is widely adopted for its capability in processing image data through effective feature extraction. To optimize these models, we employ Stochastic Gradient Descent (SGD) with a learning rate of 0.001, and each model is trained over 100 epochs. This strategy replicates real-world conditions of non-IID data distribution among network clients, which enables an assessment of the model's ability to generalize effectively.

\subsection{Multivariate Data Experiments}


In the centralized training approach, the baseline performance involves training the model with raw data points in a conventional manner, with the batch size set at 50. As illustrated in Table \ref{exp_scalar_regular_mlp}, our representative-based centralized learning method achieves the same accuracy as the baseline, demonstrating that for a small dataset, adopting the representative-based approach does not sacrifice prediction accuracy while it enhances interpretability for human understanding.
 For the representative-based distributed training over non-IID data distribution, our method is compared against FedAVG in Table \ref{exp_scalar_fed_mlp}, showing a slight performance difference of 0.95\% lower than FedAVG. However, subsequent sections will demonstrate that as the complexity of the ML model increases, our method's performance is the same with or even surpasses that of FedAVG in the distributed setting.

\begin{table}[ht]
\caption{Comparisons between different centralized learning methods on MLP model for Multivariate dataset.}
\begin{tabular}{p{2.3cm}p{2.3cm}p{2.3cm}}
\hline
Method   & Accuracy &  Batch Size \\ \hline
Baseline &    94.29\%                 &  50                          \\ \hline
    \bf{Our Method}    &   94.29\%              &  50                     \\ \hline
\end{tabular}
\label{exp_scalar_regular_mlp}
\end{table}

\begin{table}[ht]
\caption{Comparisons between different 
distributed learning methods on MLP model for Multivariate datasets.}
\begin{tabular}{p{1.6cm}p{1.6cm}p{1.6cm}p{1.6cm}}
\hline
Method   & Accuracy &  Clients  &  Batch Size  \\ \hline
    FedAVG &    92.38\%       &  2          &  50                          \\ \hline
    \bf{Our Method}    &   91.43\%     &  2         &  50                     \\ \hline
\end{tabular}
\label{exp_scalar_fed_mlp}
\end{table}

\subsection{MNIST Data for Centralized Training}



In this section, we assess our methodology by employing MLP and CNN models on the high-dimensional MNIST dataset using a centralized training approach. We compare the performance of our representative-based training against the baseline for both models, with the findings shown in Table \ref{exp_mnist_regular_mlp} and Table \ref{exp_mnist_regular_cnn}, respectively. The results indicate a correlation between model complexity and the performance of our method: for the MLP model, the representative-based method has a performance 2.25\% lower than the baseline. Conversely, with the more complex CNN model, our approach narrows the gap to just 0.13\% away from the baseline. This suggests that for high-dimensional datasets such as MNIST, the representative-based approach in centralized training aligns more closely with baseline performances as model complexity increases.

\begin{table}[ht]
\caption{Comparisons between different centralized learning methods on MLP model for MNIST dataset.}
\begin{tabular}{p{2.3cm}p{2.3cm}p{2.3cm}}
\hline
Method   & Accuracy &  Batch Size \\ \hline
Baseline &    93.97\%                 &  64                          \\ \hline
    \bf{Our Method}    &   91.72\%              &  64                     \\ \hline
\end{tabular}
\label{exp_mnist_regular_mlp}
\end{table}

\begin{table}[ht]
\caption{Comparisons between different distributed learning methods on CNN model for MNIST dataset.}
\begin{tabular}{p{2.3cm}p{2.3cm}p{2.3cm}}
\hline
Method   & Accuracy &  Batch Size \\ \hline
Baseline &    98.87\%                 &  64                          \\ \hline
    \bf{Our Method}    &   98.74\%              &  64                     \\ \hline
\end{tabular}
\label{exp_mnist_regular_cnn}
\end{table}

\subsection{MNIST Data for Distributed Training}


In this section, we assess our representative-based approach using MLP and CNN models on the MNIST dataset under a distributed learning scenario with non-IID data distribution across various clients. The performance comparisons of our method with FedAVG for both models are shown in Table \ref{exp_mnist_fed_mlp} for the MLP model and Table \ref{exp_mnist_fed_cnn} for the CNN model. We note that for the MLP model our representative-based learning achieves comparable results to FedAVG, outperforming it by 0.18\% with $K=2$ clients, yet slightly underperforming by 0.32\% when the client count increases to $K=4$.
Significantly, with the more complex CNN model, the representative method consistently surpasses FedAVG, achieving a 0.14\% improvement when $K=2$ and a notable 1.62\% increase when $K=4$.

\begin{table}[ht]
\caption{Comparisons between different distributed learning methods on MLP model for MNIST dataset.}
\begin{tabular}{p{1.6cm}p{1.6cm}p{1.6cm}p{1.6cm}}
\hline
Method   & Accuracy &  Clients  &  Batch Size  \\ \hline
FedAVG &    90.75\%       &  2          &  64                          \\ \hline
    \bf{Our Method}    &   90.93\%     &  2         &  64                     \\ \hline
    FedAVG &    90.28\%       &  4          &  64                          \\ \hline
    \bf{Our Method}    &   89.94\%     &  4         &  64                     \\ \hline
\end{tabular}
\label{exp_mnist_fed_mlp}
\end{table}

\begin{table}[ht]
\caption{Comparisons between different distributed learning methods on CNN model for MNIST dataset.}
\begin{tabular}{p{1.6cm}p{1.6cm}p{1.6cm}p{1.6cm}}
\hline
Method   & Accuracy &  Clients  &  Batch Size  \\ \hline
FedAVG &    98.18\%       &  2          &  64                          \\ \hline
    \bf{Our Method}    &   98.32\%     &  2         &  64                     \\ \hline
    FedAVG &    95.97\%       &  4          &  64                          \\ \hline
    \bf{Our Method}    &   97.59\%     &  4         &  64                     \\ \hline
\end{tabular}
\label{exp_mnist_fed_cnn}
\end{table}

\begin{figure}[ht]
\centering
\includegraphics[width=0.47\textwidth]{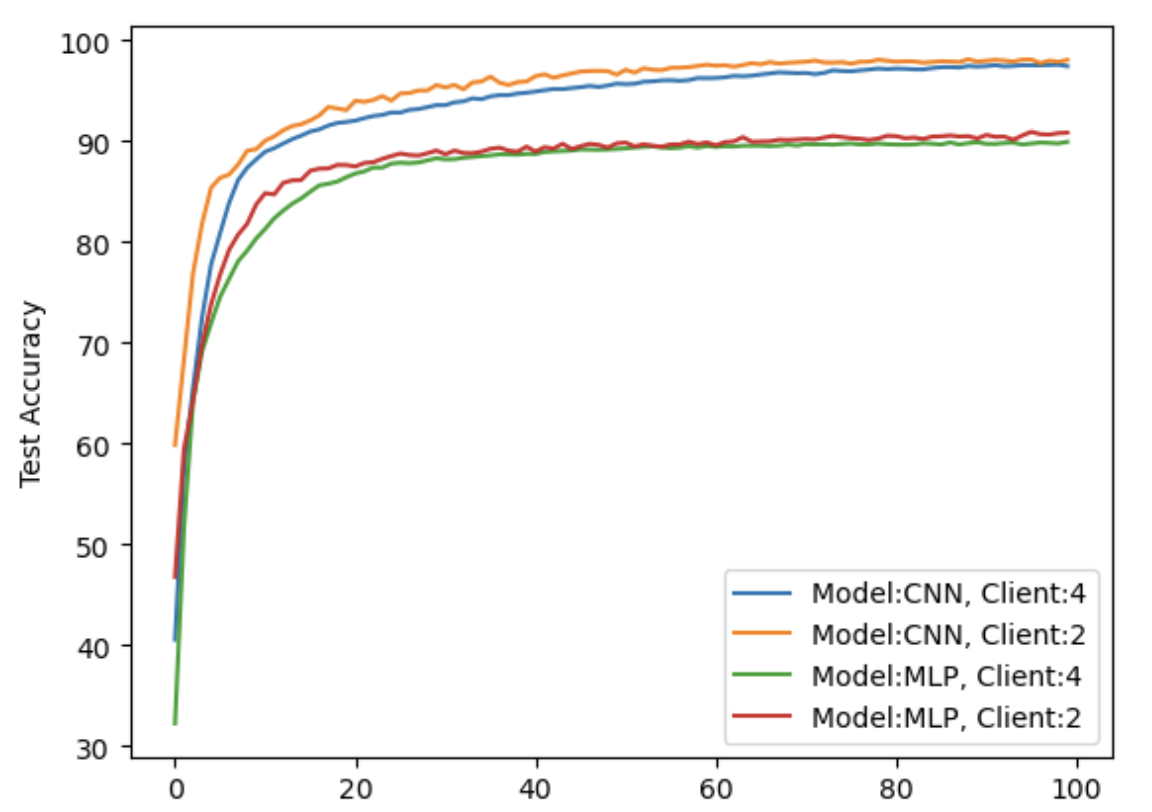} 
\caption{Convergence of representative-based distributed learning for MINST dataset.} 
\label{fig:fed_all} 
\end{figure}

The convergence patterns of representative-based distributed training for both MLP and CNN models, across varying numbers of clients, are shown in Fig. \ref{fig:fed_all}. It is observed that convergence is enhanced with a reduced number of clients for each model. Nonetheless, the convergence trajectory remains consistently smooth and fast across both models, irrespective of the client count, demonstrating no substantial decline in convergence speed as the number of clients rises from $2$ to $4$.

\begin{figure}[ht]
\centering
\includegraphics[width=0.45\textwidth]{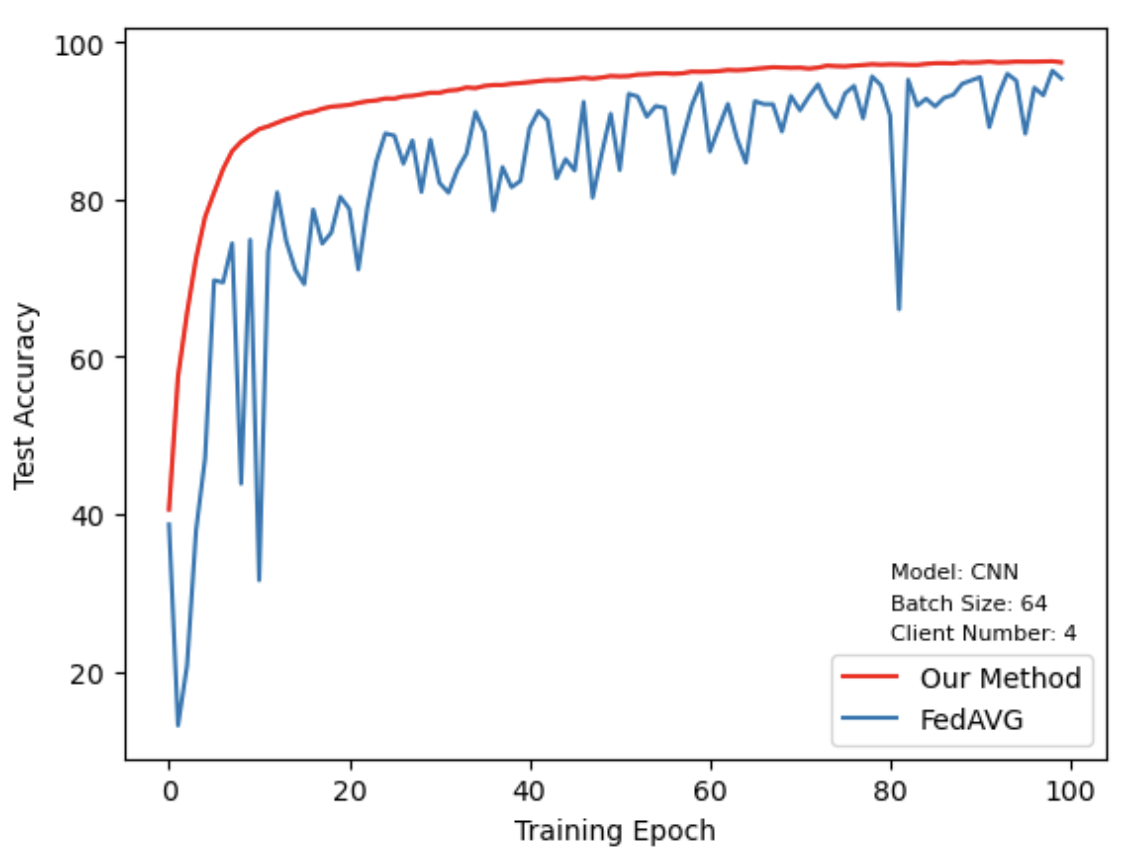} 
\caption{Comparison of convergence between representative distributed learning and FedAVG for MNIST dataset.} 
\label{fig:fed_cnn_4} 
\end{figure}

The convergence comparison between our method and FedAVG for training the CNN model is illustrated in Fig. \ref{fig:fed_cnn_4}. This comparison shows that our approach significantly outperforms FedAVG's training process in terms of both stability and convergence speed. FedAVG often exhibits poor convergence behavior due to the simple averaging of gradients from different clients, leading to considerable fluctuations in the learning curve. On the other hand, the improved accuracy, as detailed in Table \ref{exp_mnist_fed_cnn}, and the enhanced convergence observed in Fig. \ref{fig:fed_cnn_4}, can be attributed to the representative construction process. By integrating a specific magnitude of perturbations to the original dataset's average, the representative effectively introduces a form of regularization for each client. This form of regularization occurs as the server trains with ensembles of representatives from various clients in a batch, effectively minimizing the likelihood of divergence. These findings confirm that our representative-based strategy presents a viable and promising alternative to traditional distributed learning approaches.

\subsection{Improvements  Achieved by Incorporating the Residual}

\begin{figure}[ht]
\centering
\includegraphics[width=0.47\textwidth]{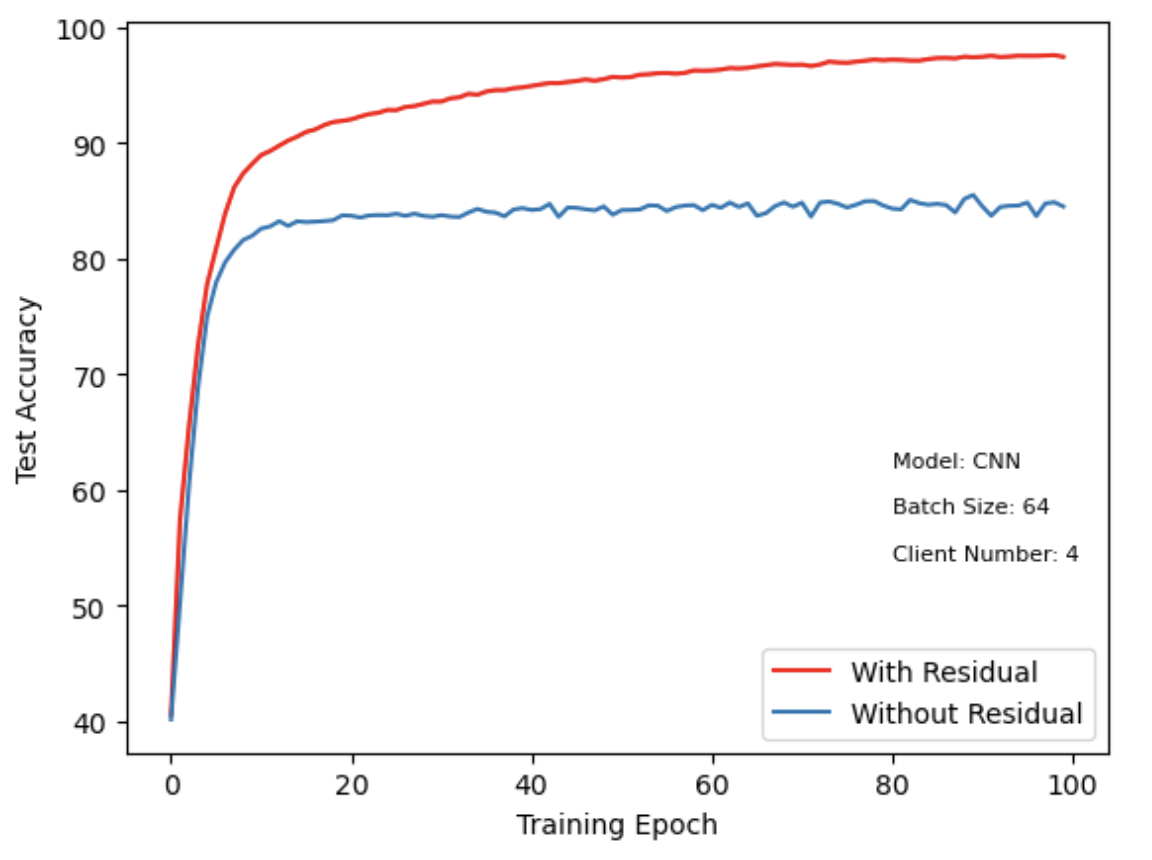} 
\caption{Convergence of representative-based distributed learning for MINST dataset with and without residual.} 
\label{fig:Residual} 
\end{figure}

In constructing the representative, we incorporate residuals to adjust for discrepancies between the gradient induced by the representative and that induced by the raw data. The positive impact of including residuals on model accuracy is illustrated in Fig. \ref{fig:Residual}, which contrasts the training convergence and accuracy of the CNN model in a distributed framework, comparing scenarios with and without residual terms. The adoption of residual corrections has notably enhanced system performance, acting as an effective counterbalance to any potential decline in representative data quality. Furthermore, this strategy helps mitigate any adverse effects or uncertainties that arise from the process of representative aggregation at the server.

\section{Disscusion}



The simulations presented above demonstrate that our method can match or surpass the accuracy and convergence rates of conventional Federated Learning (FL) techniques, such as FedAVG. This advantage becomes increasingly apparent with the rising complexity of the model and the growth in the number of clients.

This enhancement is likely due to the representative construction process, which involves introducing perturbations to the average of the raw dataset. This process effectively acts as a form of regularization, ensuring the maintenance of high-quality, informative inputs. Training the server with ensembles of these representatives helps to stabilize the learning process, in contrast to merely averaging diverse gradients. Such regularization facilitates more effective learning from representative data formats, a benefit that becomes more evident in complex models.

\begin{figure}[ht]
\centering
\includegraphics[width=0.47\textwidth]{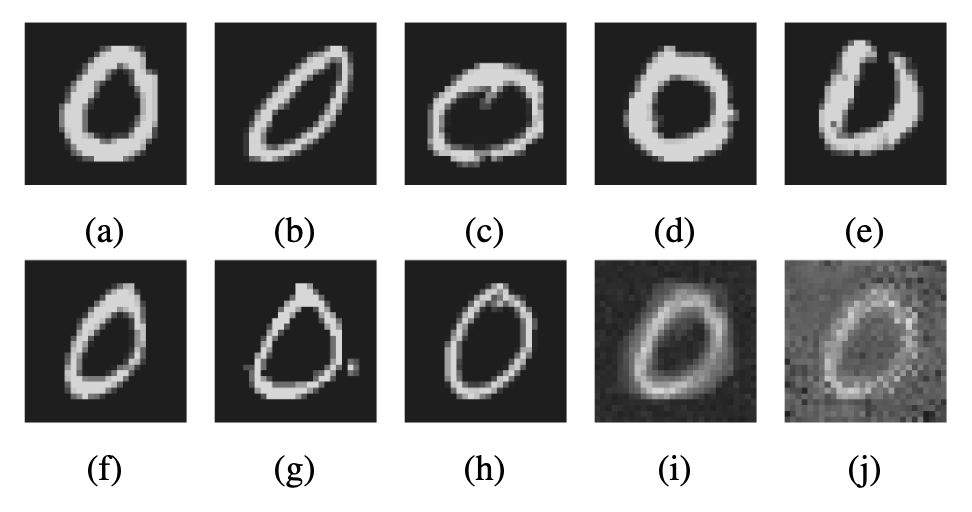} 
\caption{Representatives constructed from raw images with MLP and CNN models.} 
\label{fig:sample} 
\end{figure}

It is worth noting that one primary advantage of our approach is that it enhances the interpretability and comprehensibility of the data processing phase and improves the effectiveness of data transmission between the servers and the clients in a distributed learning setup. Unlike traditional compression or quantization methods, the representative technique allows for transmitting visually comprehensible information, facilitating a more intuitive understanding of the shared data among participants in the distributed network. In Fig. \ref{fig:sample}, we present a visualization of the representative for eight MNIST images labeled 0, ranging from (a) to (h), with (i) illustrating the representative generated via MLP and (j) via CNN. Notably, the representative from the simpler MLP model (i) appears closer to a natural image with label 0, whereas the CNN-generated representative (j) exhibits additional perturbations, particularly in the background. This observation suggests that constructing representatives with more complex models like CNNs necessitates more perturbations to the original input. This can add difficulty to human visualization and perception, indicating a trade-off between model complexity and the ease of interpreting representatives.


\section{Conclusions and Future Work}

This study introduced a novel representative-based methodology that transforms multiple data points from a client into a single virtual data point, establishing a new approach in data processing and aggregation for distributed learning systems. This method enhances privacy and communication efficiency, which are crucial benefits in distributed learning contexts. Moreover, its capacity to simplify complex, diverse datasets into digestible forms facilitates the complex dynamics of human-ML collaboration.
Simulation outcomes indicate that our approach can match or exceed the accuracy and convergence rates of traditional Federated Learning methods, particularly as model complexity increases and with the growth in the number of clients. 

Moving forward, we will expand our analysis by incorporating communication requirements and conducting more comprehensive comparisons with a wider range of architectures to validate our findings. Additionally, we will explore how human oversight and interaction can be incorporated into the representative-based learning and decision-making process, aiming to improve situational awareness and achieve a seamless integration between human intuition and machine intelligence.

\bibliographystyle{IEEEtran}
\bibliography{refer.bib}

\end{document}